\documentclass[%
 amsmath,amssymb,
 aps,
]{revtex4-2}

\usepackage[dvipdfmx]{graphicx}
\usepackage{bm}      
\usepackage{setspace}
\usepackage{amsfonts}
\usepackage{braket}

\newcommand{\pdev}[3]{{\text{d}^{#3} #1}/{\text{d}#2^{#3}}}

\newcommand{\NN}{{\mathbb N}}

\usepackage{amsmath}	
\begin{document}
\preprint{APS/123-QED}

\title{Photonic neural field on a silicon chip: large-scale, high-speed neuro-inspired computing and sensing
}

\author{Satoshi Sunada$^{1,2}$ and Atsushi Uchida$^3$
}
 \email{sunada@se.kanazawa-u.ac.jp}
\affiliation{%
$^1$Faculty of Mechanical Engineering, Institute of Science and
Engineering, Kanazawa University\\
Kakuma-machi Kanazawa, Ishikawa 920-1192, Japan \\
$^2$Japan Science and Technology Agency (JST), PRESTO, 4-1-8 Honcho,
 Kawaguchi, Saitama 332-0012, Japan\\
$^3$Department of Information and Computer Sciences, Saitama University,\\
255 Shimo-Okubo, Sakura-ku, Saitama City, Saitama, 338-8570, Japan.
}

\date{\today}

\begin{abstract}
Photonic neural networks have significant potential for high-speed neural processing with low latency and ultralow energy consumption.
However, the on-chip implementation of a large-scale neural network 
is still challenging owing to its low scalability.
Herein, we propose the concept of a {\it photonic neural field}
and implement it experimentally on a silicon chip 
to realize highly scalable neuro-inspired computing.
In contrast to existing photonic neural networks,
the photonic neural field is a spatially continuous field that nonlinearly 
responds to optical inputs, and its high spatial degrees of freedom 
allow for large-scale and high-density neural processing on a 
millimeter-scale chip. 
In this study, we use the on-chip photonic neural field 
as a reservoir of information and 
demonstrate a high-speed chaotic time-series prediction with low errors
using a training approach similar to reservoir computing.
We discuss that
the photonic neural field is potentially capable of 
executing more than one peta multiply-accumulate operations per second 
for a single input wavelength on a footprint as small as 
a few square millimeters.
In addition to processing, the photonic neural field can be 
used for rapidly sensing the temporal variation of an optical phase, 
facilitated by its high sensitivity to optical inputs. 
The merging of optical processing with optical sensing paves 
the way for an end-to-end data-driven optical sensing scheme.
\end{abstract}

\maketitle

\section{Introduction}
The recent rapid advancement of machine learning technologies has been 
driven by massive computing power, developments in 
special purpose hardware, and the availability of large datasets.
Although the accelerated increase in the volume of data in modern society 
enables the development of larger and more complex machine learning models, 
it poses considerable challenges for the present electronic computing hardware 
in terms of both computing speed and power consumption \cite{Furber_2016,6817512}.
Such issues have emerged as a major bottleneck for artificial intelligence (AI)
\cite{arXiv200705558} and motivated the development of novel AI computing hardware and concepts \cite{Furber_2016,Merolla668,Markovic:2020aa,9049105,Shastri:2021aa,Wu:2021aa,TANAKA2019100,8262616,Du:2017aa,Appeltant:2011ab}.
%
Recently, numerous studies on photonic neural networks have revealed 
their immense potential to overcome the major bottleneck of electronic 
computing hardware \cite{9049105,Shastri:2021aa,Tait:2017aa,PhysRevX.9.021032,Lin1004,9064516,Feldmann:2021aa,Xu:2021aa,Kitayama:2019,Hotchip_9220525,Shen:2017aa} and suggested the achievement of ultrahigh computing speed over the 
10-THz-wide optical telecommunication band in an ultralow 
energy-efficient manner.
Instead of using electronic devices, the utilization 
of photonic architecture allows for ultrahigh-speed linear operations, 
such as fully connected matrix operations and
convolutional operations \cite{Feldmann:2021aa,Xu:2021aa,Hotchip_9220525,Shen:2017aa}, which are essential in neural network processing, 
at the speed of light. 
Although the potential of photonics has been reported 
decades ago \cite{Goodman:78,Goodman:91}, 
the recent advancements in photonic integration 
and telecommunication technologies 
have revived interests in various types of photonic computing architectures. 

Among various photonic neural network architectures, 
photonic reservoir computing (RC)
has recently gained increasing research interest \cite{VanderSande:2017aa,Paquot:2012aa,Brunner:2013aa,Vandoorne:2014aa,Nguimdo:14,Ortin:2015aa,Vinckier:15,PhysRevX.7.011015,Takano:18,Sunada:2019aa,Uchida:2020JJAP,8807158,Paudel:20,Sunada:20,PhysRevX.10.041037,Harkhoe:20,Nakajima:2021aa,arXiv_Borghi_2021}.
Reservoir computers use networks of random connections and nonlinearities, called reservoirs.
The reservoir generates a large number of complex behaviors driven by an input, and is used to solve a specific computational problem \cite{Jaeger78,Maass:2002,VERSTRAETEN2007391}.  
As supervised training is performed only in the read-out layer for generating 
the desired outputs, the difficulty of the training in the reservoir 
and input layers can be avoided, unlike in conventional recurrent neural networks. 
Despite the simple training, high computational performance has been
 demonstrated in a series of benchmark tasks such as time-series prediction, 
wireless channel equalization, image and phoneme recognition, and 
human action recognition 
\cite{Brunner:2013aa,PhysRevX.7.011015,Nakajima:2021aa,Antonik:2019aa}. 

However, photonic implementations of reservoirs as well as standard 
neural networks have faced challenging issues. 
One of the primary issues pertains to low scalability, which prevents  
the implementation of a large-scale and high-density network 
on a photonic chip and its application in practical computational tasks. 
For instance, at most, 16 artificial neurons have been implemented 
on a silicon chip \cite{Vandoorne:2014aa}. 
Although a time-division multiplexing method
has been used to virtually create more neurons in a time domain, 
the use of the multiplexing method inevitably results in
a trade-off between the number of neurons 
and the required processing time
\cite{Takano:18,Harkhoe:20,Nakajima:2021aa,arXiv_Borghi_2021}. 
Moreover, in most photonic or optoelectronic systems, 
the computing speed is limited by the bandwidth of electric or 
optoelectronic components used for introducing nonlinear activation functions 
and feedback loops to construct a recurrent network 
\cite{8807158,Paudel:20,PhysRevX.10.041037,Antonik:2019aa}. 
Accordingly, the potential ultrawide bandwidth of photonic systems 
has not yet been completely exploited.  

In this study, we propose a novel photonic architecture 
to resolve the aforementioned issues.
The key idea is to utilize the continuous spatial degrees of 
freedom of an optical field, for example, 
a speckle field, which is generated by a complex interference 
of the guided modes 
with different phase velocities in a multimode waveguide.  
Such an optical field works as a spatially continuous neural network 
with complex internal connections, i.e., a {\it neural field}.  
We show that the use of a neural field
allows for large-scale and high-density neural processing 
owing to its high spatial degrees of freedom.
The main processing is optically performed on a tiny chip, 
and the bandwidth is not limited by electronic feedback components, 
unlike previous spatial RC systems
\cite{8807158,Paudel:20,PhysRevX.10.041037,Antonik:2019aa}. 
We discuss that the on-chip photonic neural field
is potentially capable of the processing of 
more than one peta multiply-accumulate (MAC) 
operations per second, which is a dominant factor in neural processing.
The processing rate is boosted by more than an order of magnitude 
compared to previous approaches.
Moreover, we discuss that 
the proposed photonic chip can be used not only for a processing unit 
but also for a photonic sensing unit.
The proposed chip can operate as an interface between an analog world 
and a digital domain, as it can directly process optical analog inputs. 
As a proof-of-concept for a novel data-driven optical sensing scheme, 
we demonstrate the rapid sensing of optical phase variations 
using the proposed chip.
  
\section{Basic principle \label{sec_principle}}
\begin{figure}[htbp]
\centering\includegraphics[width=8.5cm]{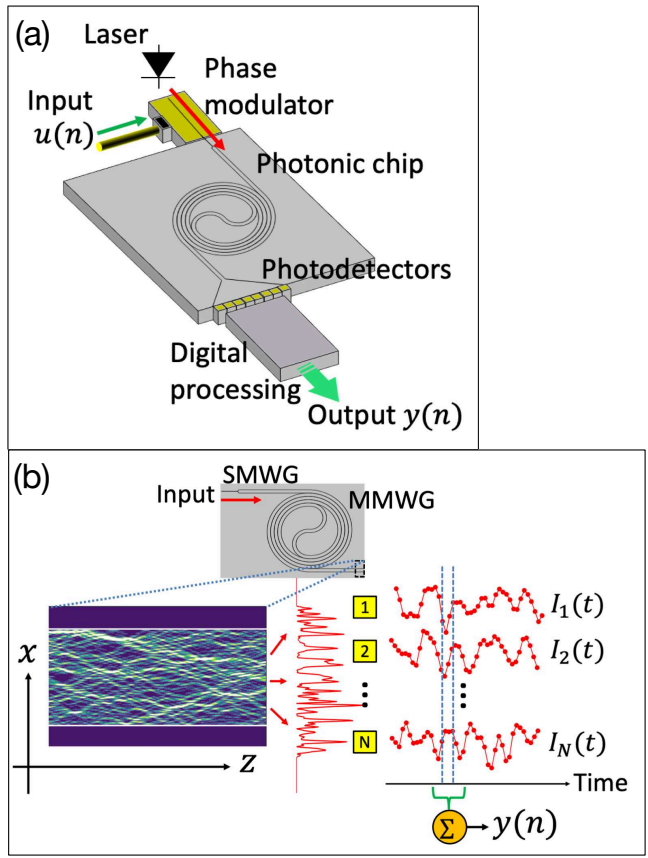}
\caption{\label{fig1}
(a) Conceptual schematic of the proposed photonic architecture.
(b) Generation of speckles as neural fields in a multimode waveguide (MMWG).
The singlemode waveguide (SMWG) is used as an input port.
The inset shows the numerical result of the intensity 
field around the end of the MMWG
with width of 25 $\mu$m and length of $39$ mm.
$I(x,t)$ is sampled at $N$ points,
and the sampled signals are used as nodes (virtual neurons) 
to compute $y(n)$. 
}
\end{figure}

Figure~\ref{fig1}(a) shows a conceptual schematic of 
the proposed photonic architecture, 
which consists of a continuous wave (CW) laser, optical phase modulator, 
a photonic chip,
and photodetectors, combined with a digital post-processing unit.  
This architecture is designed to output a signal, $y(n)$ $(n \in \NN)$, that 
responds to a time-dependent input signal, $u(n)$, 
and is trained with a training dataset, 
$\{u(n),y_{tag}(n)\}_{n=1}^{T_n}$, where $y_{tag}(n)$ is a target function. 
A major portion of the architecture 
is a multimode spiral waveguide on the photonic chip, 
which allows for the mapping of input information 
into a high-dimensional feature space, accompanied with 
a time-delay (storage) of past input information.  
The high-dimensional mapping is essential for 
facilitating classification and pattern recognition, 
whereas the storage of past inputs, i.e., a short-term memory, 
is required 
for time-dependent signal processing such as time-series prediction 
\cite{Appeltant:2011ab,Dambre:2012aa}.
Although these operations typically accompanies 
a high computational cost in existing digital computers,    
the present chip can execute
the mapping and delay (memory) processing 
through optical propagation in the multimode spiral waveguide.

For optical encoding of input signal $u(n)$, 
the laser light is phase-modulated 
with $u(t)$, where $u(t)$ holds as $u(n)$ 
for a time period of $n\tau \le t < (n+1)\tau$. 
The modulated light is transmitted to a singlemode waveguide and 
coupled to the multimode spiral waveguide, as depicted 
in Fig.~\ref{fig1}(b). 
The input field is provided 
by $E_{in}(x,t) = E_0(x)e^{i\alpha u(t)+i\omega_0 t}$, 
where $\alpha$, $\omega_0$, and $x$ denote 
the amplitude of the phase modulation, the angular frequency of 
the input light, and the transverse dimension of the propagation, 
respectively, and $E_0(x)$ is the transverse spatial profile of 
the input field. 
As the multimode waveguide is designed to support several tens of 
the guided modes with different phase velocities,  
the input field is transformed by the optical propagation 
and interference of these guided modes,
and eventually, a complex speckle field is generated 
at the end of the multimode waveguide [Fig.~\ref{fig1}(b)].  
For a multimode waveguide of length $L$, 
the output field $E_{out}(x,t;L,\omega_0)$ 
can be approximately expressed as follows \cite{Sunada:20,Saleh:1985}, 
\begin{align}
 E_{out}(x,t;L,\omega_0) \approx \sum_{m=0}^{M-1} \sum_{m'=0}^{M-1} a_{m'}(\omega_0)
\int_{-\infty}^{t} h_{mm'}(t-t',L)e^{i(\alpha u(t')+\omega_0t')}dt'\Psi_{m}(x,\omega_0),
\label{eq_out1}
\end{align}
where 
$\Psi_m(x,\omega_0)$ and $a_m(\omega_0)$ denote the transverse spatial profile of 
the $m$th guided mode ($m \in \{0,1,\cdots,M-1\}$) and 
coupling coefficient between the $m$th guided mode 
and the input field for input angular frequency $\omega_0$. 
$a_m(\omega_0)$ can be determined 
based on a boundary condition, 
e.g, $E_{in}(x,t) = E_{out}(x,t;0,\omega_0)$, 
in absence of radiation mode couplings. 
In Eq.~(\ref{eq_out1}), $|\pdev{u}{t}{}| \ll \omega_0$ 
is assumed such that $\Psi_m(x,\omega_0)$ and $a_m(\omega_0)$ 
are not affected by the phase modulation.
$h_{m,m'}(t,L)$ 
represents the response at a distance $z = L$ and time $t$ 
in $m$th mode, resulting from an impulse at the initial position $z = 0$
and time $t = 0$ in the $m'$th mode \cite{Saleh:1985}.
This coupling is attributed to the bending of the spiral waveguide 
or the scattering from the rough surface of the waveguide. 
As expressed in Eq. (\ref{eq_out1}), 
input $u(t)$ is nonlinearly transformed and 
encoded as a continuous spatial profile 
at the end of the multimode spiral waveguide.
Interestingly, the information of the past input, $u(t')$, 
for $t' < t$ is dispersed in space and time domains, 
owing to the modal dispersion of the multimode waveguide, 
i.e., the mode-dependence of the impulse response, 
$h_{mm'}(t,L)$.
Thus, the past information is available for processing time-dependent signals.

In the detection section, the output field is measured using photodetectors, 
where further nonlinear conversion is conducted
as $I(x,t) = |E_{out}(x,t;L,\omega_0)|^2$.
In this study, $I(x,t)$ is referred to as a neural field 
because it arises from the encoding of input information 
and nonlinear activation based on Eq. (\ref{eq_out1}). 
To compute output $y(n)$, 
$I(x,t)$ is sampled at position $x_i$ ($i \in \{1,2,\cdots, N\}$) 
$K$-times at a time interval $\Delta t = \tau/K$ for 
$n\tau \le t < (n+1)\tau$, where $t=n\tau$ represents the time at which
the information of input $u(n)$ reaches the end of 
the multimode waveguide.  
The output, $y(n)$, is given using $NK$-sampled neurons as
\begin{align}
 y(n) = \sum_{i=0}^{N}\sum_{k=0}^{K-1}
w_{i,k}I_i(n\tau + k\Delta t), \label{eq_out2}
\end{align}
where $I_i(t) = I(x_i,t)$ for $1 \le i \le N$ and $I_0(t) = 1$ 
is a constant for bias.
In Eq. (\ref{eq_out2}), the weight, $w_{i,k}$, 
is trained such that output $y(n)$ corresponds 
to target $y_{tag}(n)$ for any $n \in \{1, \cdots, T_n\}$.
The training is performed using a simple ridge regression to minimize the error 
$\sum_{n=1}^{T_n}|y_{tag}(n)-y(n)|^2 + \gamma \sum_{i,k}w_{i,k}^2$, where
$\gamma$ is a regularization parameter used for averting 
an ill-conditioned problem. 

A benefit of the photonic neural field is that it allows for 
using a large number of neurons for computing $y(n)$ 
through spatial sampling.
In particular, wavelength-scale sampling enables high-density 
neural processing for obtaining $y(n)$ on a tiny chip, 
as shown later. 
In addition, 
the entire photonic system shown in Fig.~\ref{fig1}(a)
does not utilize a feedback loop to construct a recurrent network. 
As mentioned earlier, the modal dispersion of the multimode waveguide 
enables us to use the information of past inputs;
thus, the proposed photonic system can be much simpler than 
existing photonic systems with feedback loops 
\cite{Paudel:20,PhysRevX.10.041037}, and 
the computing speed is not limited by the feedback loop.

We remark that impulse $h_{mm'}$ and wave function $\Psi_m$ in 
Eq.~(\ref{eq_out1}) are dependent on input angular frequency $\omega_0$
(input wavelength $\lambda_0 = 2\pi/\omega_0$), 
and thus, various spatiotemporal behaviors can be generated 
using lights of different wavelengths in parallel.  
The parallel computing with spatiotemporal dynamics based on 
wavelength-division multiplexing is discussed in Sec. \ref{sec6}.  

\section{Methods}
\subsection{Photonic chip}
Figure \ref{fig2}(a) shows a spiral-shaped multimode waveguide 
fabricated on a silicon chip. 
The 220 nm thick silicon layer was etched to form the multimode 
spiral waveguide. 
The width and length of the multimode waveguide are 25 $\mu$m and 39 mm, 
respectively. 
%
%
The multimode waveguide is bended in form of
two Archimedean-shaped spiral geometries, which are connected at the center 
by an S-bend waveguide with 300 $\mu$m radius
and coupled to the 500-nm-wide singlemode waveguide used as the input port. 
A spot-size converter was used for coupling between the singlemode waveguide
and an optical fiber. 
Consequently, the light from the input port propagates inward 
through one spiral and then outward 
through the other spiral to the output port. 
The bending of the waveguide can induce the modal coupling caused by
the propagation \cite{Redding:16}.
The distance between the adjacent arms is designed 
to be under 500 nm, which may induce a partial back-and-forth coupling 
to lengthen the optical path \cite{Redding:16}
and enhance the sensitivity to the modulation of the input light. 

\subsection{Experimental setup \label{sec_expset}}
\begin{figure}[htbp]
\centering\includegraphics[width=8.5cm]{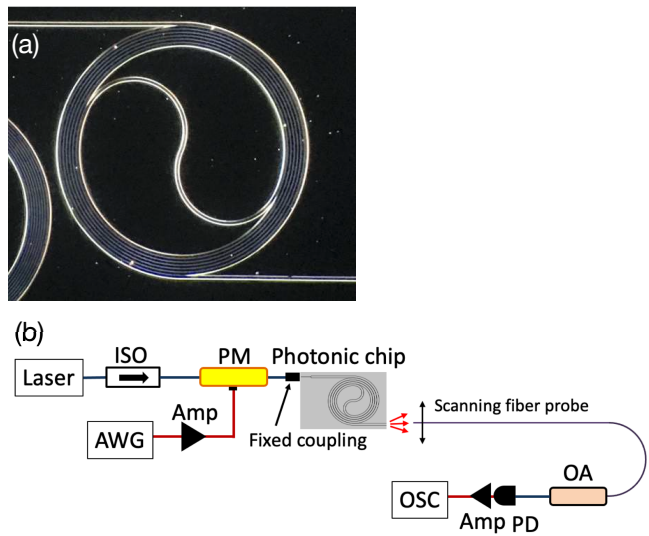}
\caption{\label{fig2}
(a) Top view of the fabricated silicon multimode spiral waveguide.
The width and length of the multimode waveguide are 25 $\mu$m and 39 mm, 
respectively.
The radius of the S-bent waveguide at the center is 300 $\mu$m.
(b) Experimental setup for testing the performance of the photonic chip.
ISO, optical isolator; Amp, electric amplifier; AWG, arbitrary waveform
generator; PM, phase modulator; OA, optical amplifier; PD, photodetector; OSC,
digital oscilloscope.
A polarization-maintaining single-mode fiber was coupled to the photonic 
chip. 
The fiber probe (lensed fiber) was adjusted to scan the output field 
from the chip.
}
\end{figure}

Figure \ref{fig2}(b) shows the experimental setup 
used in this study to test the device performance.
A narrow-linewidth tunable laser (Alnair Labs, TLG-220, 100 kHz linewidth, 20 mW) was used as a light source.
The laser light was phase-modulated using a lithium niobate phase modulator (EO Space, PM-5S5-20-PFA-PFA-UV-UL, 16 GHz bandwidth, RF $V_\pi \approx 3.3$ V) to encode input signal $u(t)$, which was generated using an arbitrary waveform generator (Tektronix, AWG70002A, 25 Gigasamples/s (GS/s)). 
The modulated light was transmitted from a polarization-maintaining singlemode fiber to the input port of the photonic chip via a spot-size converter. 
In the initial prototype, we avoided the complication of integrating 
photodetector arrays as depicted in Fig. \ref{fig1}(a)
and measured the time evolution of the output field from the photonic chip 
following the method reported in Ref. \cite{Sunada:20} that 
is based on the scanning of the field using a lensed fiber probe.
The signal detected using the lensed fiber probe 
(with a working distance of $\sim 10 \mu$m and 
spot size of $\sim 2 \mu$m) is transmitted to a photodetector 
(New Port, 1554-B, 12 GHz bandwidth). 
The measured signal was digitized with an AC coupling mode 
of a digital oscilloscope (Tektronix, DPO72504DX, 25 GHz bandwidth, 50 GS/s) 
and post-processed with a computer.  

\section{Photonic processing}
\subsection{Time-series prediction \label{sec_time}}
As a demonstration of processing with the photonic chip, we performed 
the Santa Fe time-series prediction task \cite{298828}, 
the goal of which is to predict 
a one-step ahead of the chaotic data generated from a far-infrared laser. 
In this task, input signal $u(n)$ corresponds to the $n$th sampling 
point of the chaotic waveform, and target $y_{tag}(n)$ was set as 
the $n+1$th sampling point, $u(n+1)$.
The prediction error was evaluated using the normalized mean square error (NMSE)\cite{Kuriki:18}.
In this study, we used $T_n = 3000$ step points for training 
and 1000 step points for testing. 

\begin{figure}[htbp]
\centering\includegraphics[width=8.5cm]{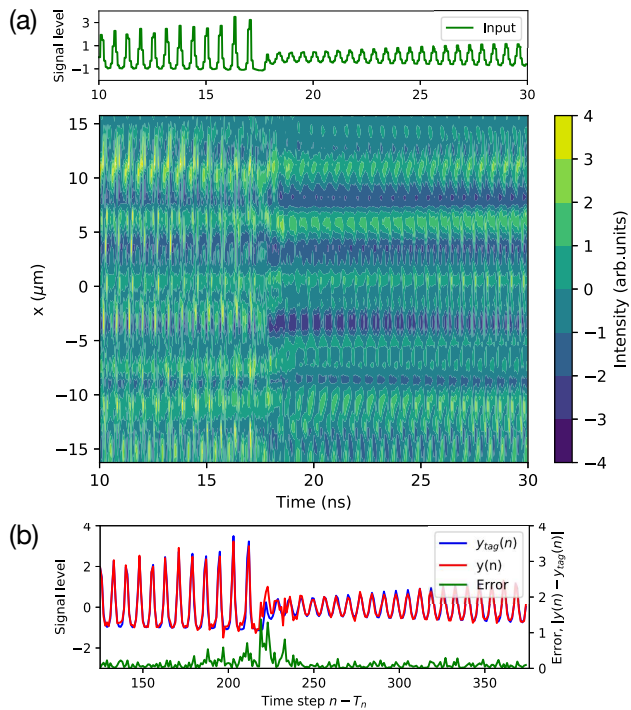}
\caption{\label{fig3}
(a) Input signal $u(t)$ as a function of time $t = (n - T_n)\tau$ in the chaotic time-series prediction task (upper panel).
$T_n = 3000$. 
Input rate $1/\tau$ was set as 12.5 GS/s ($\tau = 0.08$ ns). 
The lower panel depicts the spatiotemporal evolution of the neural field
$I(x,t)$ at the end of the waveguide. 
For the purpose of displaying, the DC component of the measured neural field
was removed, and the origin of time ($t = 0$) was adjusted 
for input signal $u(t)$. 
$x$ denotes the transversal dimension (see Fig.~\ref{fig2}).
(b) Output $y(n)$ and target $y_{tag}(n)$. 
Large prediction errors occurred after the sudden variation 
in the input signal level.
}
\end{figure}

Figure \ref{fig3}(a) shows the spatiotemporal evolution of 
the neural field $I(x,t)$ 
responding to the input light of the wavelength $\lambda = 1550$ nm 
that was modulated using the chaotic signal, $u(n)$, 
at an input rate of $1/\tau = 12.5$ GS/s ($\tau = 0.08$ ns). 
$I(x,t)$ was measured at a spacing 
of $\Delta x \approx 500$ nm 
and time interval of $\Delta t = 0.02$ ns, 
where $x$ is the spatial coordinate 
on the waveguide cross-section at the output port. 
As portrayed in Fig.~\ref{fig3}(a), a variety of temporal responses 
to the input signal can be observed, depending on the observation 
position, $x$.
The spatiotemporal behavior results from a complex interference among 
the guided modes with various phase velocities [Eq.~(\ref{eq_out1})]. 
A typical length of the spatial correlation 
was estimated as $\sim 1.9$ $\mu$m, (see Appendix \ref{app_sec1},) 
which is comparable to the limitation of the resolution, i.e, 
the spot size ($\sim 2$ $\mu$m) 
of the lensed fiber probe used in this experiment. 
We used $N = 65$ and $K = \tau/\Delta t = 4$ sampled signals
as virtual neurons to compute output $y(n)$.
Figure~\ref{fig3}(b) presents output $y(n)$ obtained
from the virtual neurons with the total number of $NK = 260$ 
using Eq.~(\ref{eq_out2}). 
The NMSE of the prediction was approximately 0.039, 
which is superior to those 
reported in previous photonic systems (for instance, NMSEs of 0.106, 
0.109, and 0.09 have been reported in Refs.~\cite{Brunner:2013aa}, 
\cite{Takano:18}, and \cite{Sunada:20}, respectively). 
In addition, the rate of processing using the neural field 
(12.5 GS/s in our case) is much higher than the previous approaches.
The bandwidth is not limited by the electronic components and lasers
used to construct a reservoir system. 
The excellent performance in terms of prediction accuracy and processing rate 
is based on the spatial parallelism realized in the proposed photonic chip.

Moreover, the proposed photonic chip
does not contain any optical feedback loop for memory of past input.
As mentioned in Sec.~\ref{sec_principle}, 
the modal dispersion of the multimode waveguide results in
the dispersion of the information of past inputs in a temporal domain.
Information dispersion virtually operates
as a fading memory, and this partly allows for the storage of 
information related to past inputs. 
Based on a pulse-response measurement and memory capacity test, 
we estimated the time span of the short-term memory 
(information dispersion) of our multimode waveguide
as 200 ps, which
suggest that the information of the past inputs prior to
$200/\tau = $ 2.5 steps is available for the time-series processing
at a rate of $1/\tau = 12.5$ GS/s.
(See Appendix \ref{app_sec2} for the details.)  
The memory of the past 2.5 steps can contribute to the prediction performance. 
The information dispersion resulting in the memory effect can be
controlled by varying the number of the guided modes 
in the multimode waveguide and rate $1/\tau$,
depending on the tasks requiring memory.

\subsection{Wavelength multiplexing \label{sec6}}
\begin{figure}[htbp]
\centering\includegraphics[width=8.5cm]{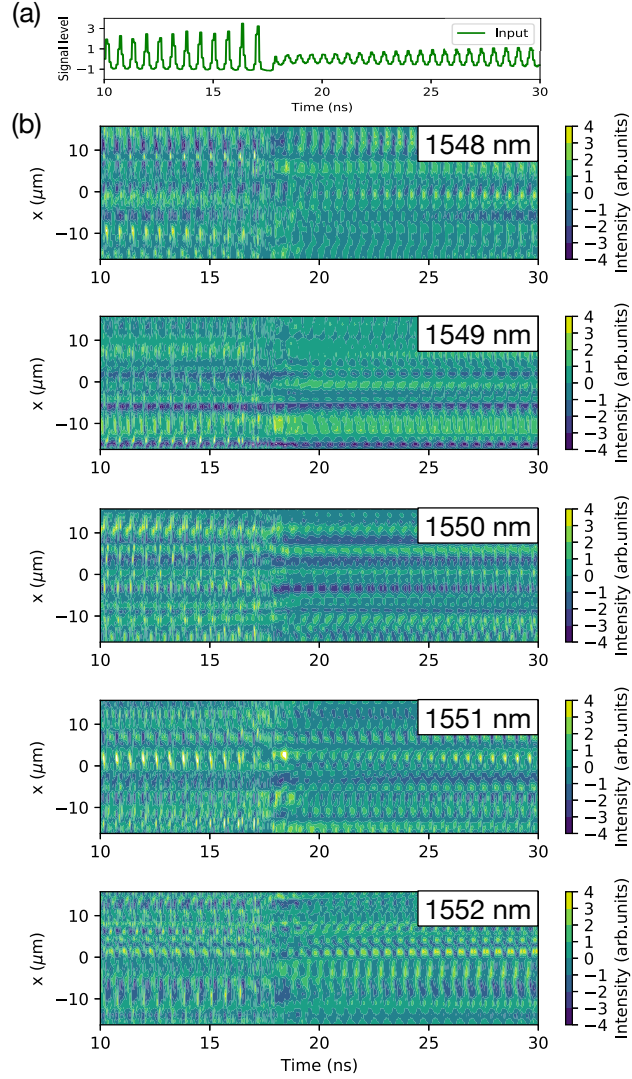}
\caption{\label{fig4}
(a) Input signal $u(t)$ as a function of time $t = (n-T_n)\tau$ in the chaotic time-series prediction task. 
Input rate $1/\tau$ was set to 12.5 GS/s. 
(b) Spatiotemporal evolution of $I^{\lambda_l}(x,t)$. $\lambda_l = 1547$ nm $+ l$ nm ($l \in \{1,2,3,4,5\}$). 
For a display purpose, the DC-component of the measured neural field
was removed, and the origin of time ($t = 0$) was adjusted 
for input signal $u(t)$. 
}
\end{figure}

\begin{figure}[htbp]
\centering\includegraphics[width=8.5cm]{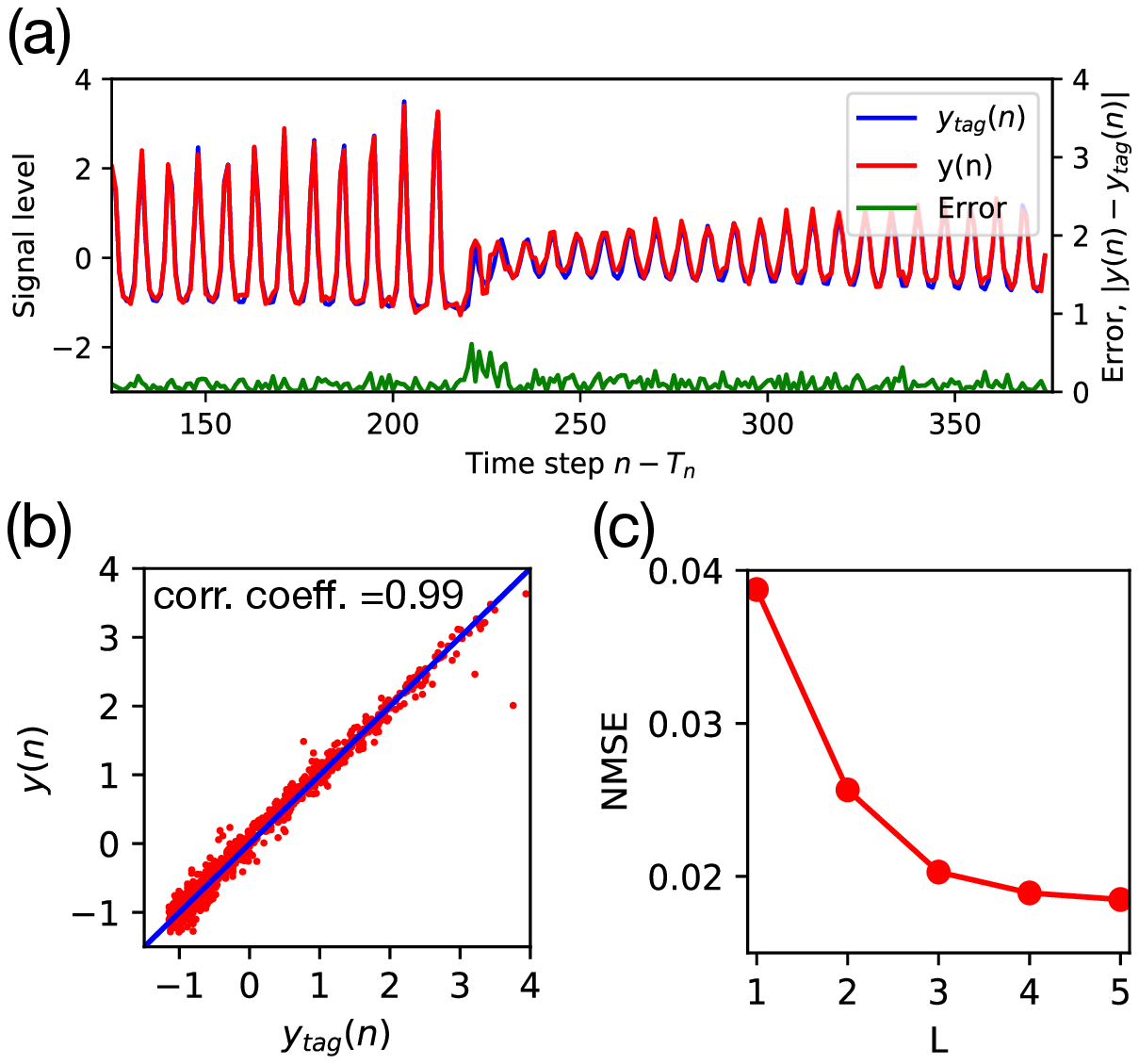}
\caption{\label{fig5}
(a) Output $y(n)$ obtained using $I^{\lambda_l}(x,t)$ ($l \in \{1,\cdots, L\}$) shown in Fig. \ref{fig4}(b).
The number of the neurons corresponds to $NKL = 1300$, 
where $N = 65$, $K = 4$, and $L = 5$. 
In addition, target $y_{tag}(n)$ is displayed for comparison. 
As compared to Fig.~\ref{fig3}(b), prediction errors are suppressed.
(b) Correlation between $y(n)$ and $y_{tag}(n)$. 
The correlation coefficient evaluated using 1000 datapoints 
was over 0.99. 
(c) The NMSE as a function of $L$.
}
\end{figure}

The proposed chip can exhibit a variety of spatiotemporal responses 
depending on the wavelength of the input light
in a {\it single} multimode waveguide and 
can utilize them as additional computational resources. 
Figure~\ref{fig4} shows the temporal evolution of the neural field
$I^{\lambda_l}(x,t)$ responding 
to input light of various wavelengths 
$\lambda_l$ ($l \in \{1,2,\cdots,5\}$).  
Each input light was phase-modulated using the same chaotic waveform signal, 
$u(n)$. 
The mean absolute value of the correlation 
between the two distinct spatiotemporal responses, 
$I^{\lambda_l}(x,t)$ and $I^{\lambda_{l'}}(x,t)$, was approximately 0.3. 
Further sensitivity to the wavelength, i.e., a lower correlation, 
can be achieved by lengthening the multimode spiral waveguide,
as demonstrated in multimode speckle spectrometers \cite{Redding:16}.  
We used the neural fields that were multiplexed as the wavelength-divisions
with an interval of $\Delta \lambda \approx 1$ nm for more accurate 
prediction. 
Output $y(n)$ was computed as 
$\sum_{k=0}^{K-1}\sum_{i=1}^N\sum_{l=1}^{L}w_{k,i,l}I^{\lambda_l}(x_i,n\tau+k\Delta t)$. 
The computation result is presented in Fig.~\ref{fig5}(a), 
where $NKL = 1300$ neurons were used for the computation 
and a significant improvement of the prediction was achieved.
The NMSE was 0.018, which is a half of that obtained from 
a single-wavelength neural field, and 
the correlation between the output $y(n)$ and 
target $y_{tag}(n)$ computed for 1000 
testing datapoints was evaluated as over 0.99 [Fig.~\ref{fig5}(b)].
Figure~\ref{fig5}(c) shows that the NMSE decreases
as the number of wavelength division multiplexing 
$L$ (the number of the neurons) increases. 
Therefore,
the photonic chip allows for multidimensional multiplexing methods 
based on wavelength-division as well as the spatial and time-divisions 
in parallel and enables computation for better predictions.  
The implementation can be achieved 
using optical filters and photodetector array \cite{Sunada:20}.

\subsection{Discussion on photonic computation}
Here, we discuss the operation of the proposed photonic chip 
from a computational perspective. 
The computations of the proposed photonic chip originates from 
high-dimensional and nonlinear mapping accompanied by the 
information of the past inputs 
onto spatial profiles of the output field, according to 
Eq.~(\ref{eq_out1}). 
Although such a high-dimensional mapping, i.e., the matrix operation 
implied from Eq.~(\ref{eq_out1}), generally involves a high computational cost, 
the processing in the proposed photonic chip
is naturally performed via
the optical propagation in the multimode waveguide with 
the energy (20 mW) of the input light. 
To evaluate the photonic chip as a computing device, 
we here use the index of the multiply-accumulate per second speed, 
which has been widely used for evaluating the performance of 
neuromorphic computers \cite{Nakajima:2021aa,Nahmias:2020,Angelina:2020}.

For a simple discussion, we here omit the operation of a modal coupling caused by the waveguide bending. 
In this case, $h_{mm'}(t,L) = \delta_{mm'}h_{m}(t,L)$ in Eq. (\ref{eq_out1}).  
Then, let $\omega_0$ and $\beta_m(\omega)$ be an angular frequency 
of the input light and the propagation constant of 
the $m$th guided mode for the angular frequency $\omega$, respectively. 
Assuming that the bandwidth of the phase-modulated input is so narrow that 
$\beta_m(\omega)$ is approximated as $\beta_m(\omega) = \beta_m(\omega_0) + \pdev{\beta_m}{\omega}{}(\omega-\omega_0) + O(|\omega-\omega_0|^2)$, 
$h_m(t,L)$ is expressed as $\delta(t-t^{g}_m)e^{-i(\beta_m^{0}L+\omega_0t^{g}_m)}$, where 
$t^{g}_m = (\pdev{\beta_m}{\omega}{})L$ 
and 
$\beta_m^{0} = \beta_m(\omega_0)$ represent 
the group delay time and propagation constant of the $m$th guided mode
for angular frequency $\omega_0$, respectively \cite{Saleh:1985}.
Accordingly, Eq.~(\ref{eq_out1}) is simplified as 
$E_{out}(x,t;L,\omega_0) \approx \sum_{m=0}^{M-1}a_{m}(\omega_0)\Psi_m(x,\omega_0)e^{i\theta(t)}$, 
where $\theta(t) = \omega_0(t-t_m) + i\alpha u(t-t^{g}_m)$ 
and $t_m = (\beta_m^{0}/\omega_0)L$.  
For $N$ spatial sampling of output field $E_{out}$, 
the above expression can be rewritten as a matrix form, 
$E_i(t) = E_{out}(x_i,t) = e^{i\omega_0 t}\sum_{m=0}^{M-1}A_{im}e^{i\alpha u(t-t^{g}_m)}$, $(i \in \{1,\cdots, N\})$, 
where the matrix element, $A_{im}=a_{m}(\omega_0)\Psi_m(x_i,\omega_0)e^{-i\omega_0t_m}$, is determined from the multimode waveguide structure.
We presume that the matrix operation is executed on a complex space,  
expecting that the photonic chip performs $6NM$ times multiplication and $2(M-1)$ times summations for time interval $\Delta t$, 
where six times and two times are required for a complex multiplication 
and summation operations, respectively \cite{Nakajima:2021aa}.  
These operations are executed $K$ times for each time step, and
the MAC/s can be evaluated as $[6NM+2(M-1)]K/\tau = 2(3NM+M-1)/\Delta t$. 
In our experiment presented in Sec.~\ref{sec_time}, 
$N = 65$ and $\Delta t = 0.02$ ns were set.
$M$ was numerically estimated as 68, as shown in Appendix \ref{app_sec3}. 
Accordingly, we consider that 
the operation corresponds to 1.333$\times 10^{15}$ MAC/s = 1.3 P MAC/s. 
The footprint of the multimode spiral waveguide is in 2.4$\times$2.0 mm$^2$.
Therefore, the MAC operations per second and per unit area 
is considered to be 278 T MAC/s/mm$^2$, which is superior than 
the other photonic reservoir chips, for instance, 
Ref.~\cite{Nakajima:2021aa} have reported 
21.12 T MAC/s using 47$\times$28 mm photonic reservoir 
chip (0.016 T MAC/s/mm$^2$).
The high performance of the proposed chip 
is based on the high-density implementation of virtual neurons,
which is enabled by wavelength-scale sampling of
the photonic neural field. 
Further improvement is enabled by increasing the waveguide width to accommodate more guided modes and parallel computations based on the wavelength-division multiplexing, as demonstrated in the previous subsection. 
In addition, replacing the multimode spiral waveguide with a microcavity 
may enable higher density implementation \cite{Sunada:2019aa}. 
Note that the ultrahigh-speed computation 
with the proposed chip can be achieved only for the fixed matrix operation, 
$A_{im}$. 
Nevertheless, the low error performance was demonstrated 
for a standard benchmark, such as a time-series prediction task, 
based on the RC-like training approach.   
Although the flexible change 
of the internal network, $A_{im}$, is challenging in the present chip, 
the installation of the Mach--Zehnder interferometer units 
presented in Ref.~\cite{Shen:2017aa} 
may enable more flexible formation of 
the internal network with a training approach based on optimal control 
\cite{PhysRevApplied.15.034092}. 

\section{Merging of sensing and processing: Sensing of optical phase variation}
One of the remarkable features of the present photonic chip 
is its ability to directly process optical analog inputs. 
In other words, it can work as an interface between an analog world and
a digital domain and can be used not only
for a processing unit but also a photonic sensing unit. 
This is based on the response sensitivities of the multimode waveguide to 
the optical phase, wavelength, and polarization of input light.
Here, as a proof-of-concept demonstration, 
we show that the proposed chip can actually
be used as a sensing unit to detect a temporal variation of the optical phase 
of an input light. 

The measurement of optical phases is fundamental 
for a variety of fields ranging from precise measurements 
for determining refractive index variations, holography, 
nondestructive testing to coherent optical communications
\cite{Frantz:79,Li:09}. 
However, the phase of a complex optical wave cannot directly be measured; 
thus, complex measurement systems based on interferometers and complex
post processing have generally been required. 
In contrast, the proposed chip based on the multimode waveguide 
encodes the variation of the optical phases as a spatial profile of 
the neural field, and the optical phase can be measured 
without any reference light, local oscillators, 
and complex post-processing.

The present experiment aims to measure the optical phase variation 
of the modulated input light.
In the experiment, the optical phase (denoted by $\phi(t)$) 
of the input light was modulated with a
pseudorandom sequence at a rate of $1/\tau = 12.5$ GS/s.
The maximum modulation amplitude was estimated as $\phi_{max} \approx 1.2\pi$. 
The modulated light was input to the multimode spiral waveguide,
and the neural field was sampled at a time interval of $\Delta t = 0.02$ ns
around $t = n\tau$. 
The experimental setup is similar with that presented in Sec.~\ref{sec_expset}. 
The target, $y_{tag}(n)$, was set as the phase variation,
$\Delta\phi(n) = \phi(n\tau)-\phi((n-1)\tau)$, at time $t = n\tau$ 
because the multimode waveguide is sensitive to the phase variation
\cite{Rawson:80}. 
The sampled signals, $\{I_i(n\tau+k\Delta t)\}_{i=1,k=0}^{N,K-1}$, 
were used to experimentally compute output $y(n)$ 
using Eq.~(\ref{eq_out2}) with $N = 65$ and $K = 4$.
The weight, $w_{i,k}$, in Eq.~(\ref{eq_out2}) 
was trained such that $y(n)$ corresponds to the phase variation, $y_{tag}(n)$.  
Using the inferred phase variation, $y(n)$, the optical phase, 
$\phi(n\tau)$, can be estimated as 
$\sum_{n'=0}^{n}y(n')$ for a sufficiently small error accumulation. 
In this experiment, $T_n = 3000$ datapoints were used for training, 
whereas 1000 points were used for the test.

\begin{figure}[htbp]
\centering\includegraphics[width=8.5cm]{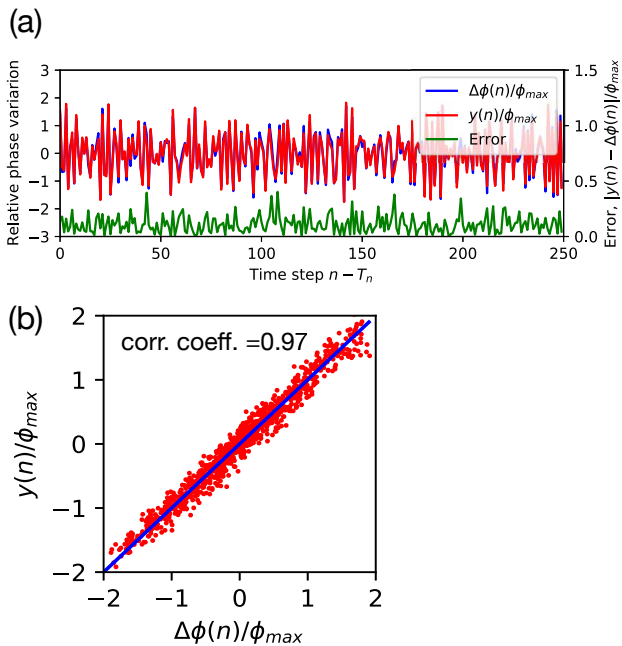}
\caption{\label{fig6}
(a) Inferred phase variation $y(n)$; 
the input light is modulated at $1/\tau = 12.5$ GS/s. 
The root-mean-square error was 0.14$\phi_{max}$.
(b) Correlation between $y(n)$ and $\Delta\phi(n)$. 
}
\end{figure}

Figure \ref{fig6}(a) and \ref{fig6}(b) show 
the inferred optical phase variation $y(n)$ and 
correlation between $y(n)$ and $\Delta\phi(n)$, respectively.  
The correlation coefficient was 0.97, and the NMSE was 0.029, 
indicating the root-mean-square error of 0.14$\phi_{max}$.

\begin{figure}[htbp]
\centering\includegraphics[width=8.5cm]{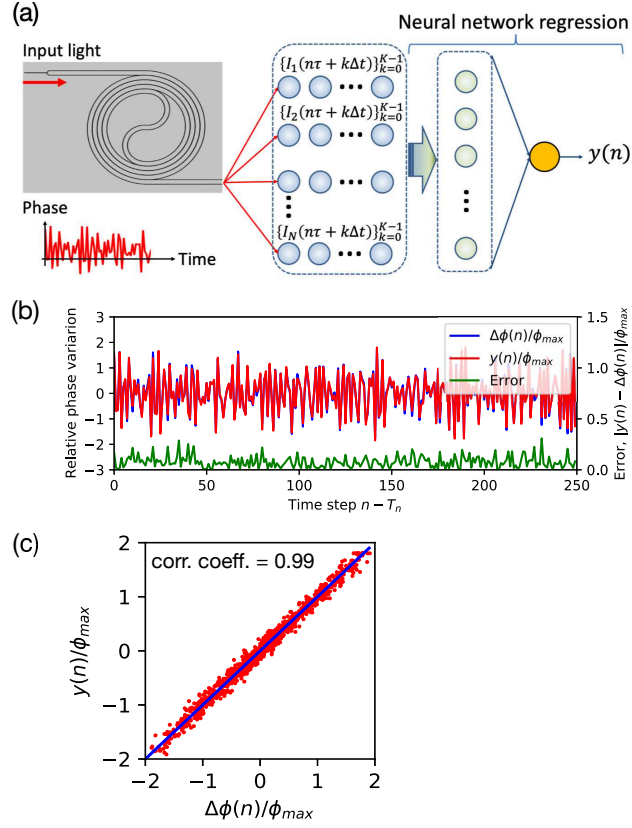}
\caption{\label{fig7}
(a) Inference of the optical phase variation of the input light 
using a single-layer feedforward neural network as a post-processor. 
The neural network consists of 150-neurons with the ReLu activation function and is trained 
such that output $y(n)$ corresponds to the input phase variation, 
$\Delta\phi(n)$, at the $n$th time step. 
(b) Inferred phase variation $y(n)$. 
The root-mean-square error was 0.09$\phi_{max}$.
(c) Correlation between $y(n)$ and $\Delta\phi(n)$. 
}
\end{figure}

To enhance the expressivity of the measurement architecture and 
further improve the measurement performance, we utilized
a neural network as a post processor, as shown in Fig.~\ref{fig7}(a). 
As a simple instance, we used a single-layer feedforward neural network, 
which consists of 150 neurons with rectified linear (ReLu) 
activation functions.  
The sampled signals, $\{I_i(n\tau+k\Delta t)\}_{i=1,k=0}^{N,K-1}$, were 
input to the neural network, and 
the network was trained such that output $y(n)$ corresponds to 
$\Delta\phi(n\tau)$.
For training, we used the Limited-memory Broyden-Fletcher-Goldfarb-Shanno (LBFGS) method \cite{LBFGS:2014}, 
which is a limited memory quasi-Newton method and enables fast 
convergence for small datasets.
Figure~\ref{fig7}(b) and \ref{fig7}(c) display
network output $y(n)$ and correlation between 
$y(n)$ and $\Delta\phi(n)$, respectively.  
The correlation coefficient was improved to 0.99, 
and the NMSE was reduced to 0.013, 
indicating the root-mean-square error of 
0.09$\phi_{max}$. 
The similar performance was confirmed when the number of the neurons 
is lesser than the number of the features, $NK$, suggesting 
that the neural network works as an appropriate 
feature extractor.

The multimode waveguide is sensitive to optical wavelength, polarization, 
and amplitude in addition to optical phase variation.
Thus, these can be simultaneously measured with an appropriate 
training approach. 
This may create an alternative possibility of multi-modal and data-driven
optical sensing.  


\section{Conclusion}
In this study, we introduced the concept of a photonic neural field 
and reported its on-chip implementation for high-density, high-speed, 
neuro-inspired photonic computing systems with a simple 
configuration.
The proposed photonic chip with a multimode spiral waveguide structure
can generate a speckle field owing to the complex interference 
of the guided modes.
The speckle field can be used as a photonic neural field because 
its continuous spatial degrees of freedom of the field 
allow for neural processing, which regards 
high-dimensional and nonlinear mapping of inputs.
The processing is optically performed and enables an excellent performance 
in chaotic time-series prediction with low errors 
at a high rate of 12.5 GS/s on a millimeter-scale chip. 
The potential operation speed is considered to correspond to
more than one peta MAC/s in a small area of 2.4$\times$2.0 mm$^2$.  
In the future, faster information processing is possible by combining 
wavelength-division multiplexing in parallel.
The use of the wavelength-division multiplexing
also enables parallel multiple processing in a single chip 
owing to the passivity of the photonic chip \cite{Sunada:20}.

In addition to optical processing, 
the proposed photonic chip can be used as a photonic sensing unit.
We demonstrated the rapid sensing of the optical phase variations 
using the proposed photonic chip.
Additional optical quantities, such as 
wavelength, polarization, and amplitude,
can be detected in parallel because the multimode speckles are sensitive to
them.
Based on the combination of sensing and processing,
the proposed approach opens a novel pathway toward data-driven, model-free, and multimodal sensing.  

\begin{acknowledgements}
This work was supported, in part, by JSPS KAKENHI (Grant
No. 19H00868, 20H042655), JST PRESTO (Grant No. JPMJPR19M4),
and Okawa Foundation for Information and Telecommunications.
The author S.S. thanks Takeo Maruyama and Takehiro Fukushima 
for useful discussion on the silicon chip fabrication. 
\end{acknowledgements}

\appendix
\section{Spatial correlation \label{app_sec1}}
\begin{figure}[h]
\begin{center}
\raisebox{0.0cm}{\includegraphics[width=12cm]{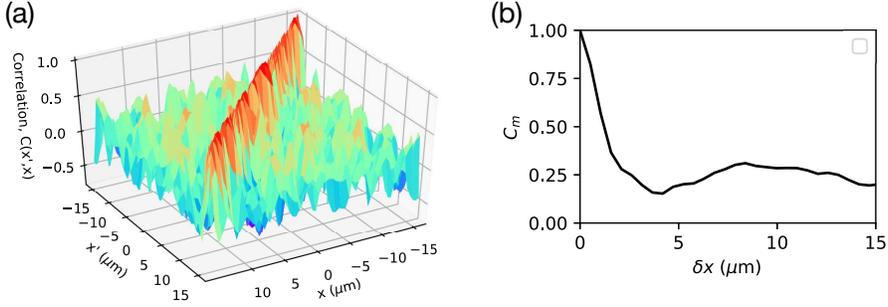}} \\
\end{center}
\caption{\label{fig_ap1}
(a) Correlation matrix $C(x,x')$. (b) Mean correlation $C_m$ as a function of $\delta x =|x'-x|$. 
}
\end{figure}

We measured the spatial correlation between two signals, 
$I(x,t)$ and $I(x',t)$, measured at
sampling position $x$ at time $t$, which is 
defined as follows:
\begin{align}
C(x,x')=\dfrac{
\langle [I(x,t)-\bar{I}_x]
[I(x',t)-\bar{I}_{x'}]
\rangle_T
}
{
\sigma_x\sigma_x'},
\end{align}
where $\langle f \rangle_T = 1/T \int^T_0 f dt$ denotes 
the mean of $f$ over measurement time $T$.
In addition, $\bar{I}_x$ and $\sigma_x$ are the mean and standard
deviation of $I(x,t)$, respectively.   
The correlation $C(x,x')$ decayed for a large $\delta x =|x'-x|$, as
shown in Fig. \ref{fig_ap1}(a).
To clarify this, we computed the mean absolute correlation value 
$C_m(\delta x) = 1/D\int |C(x,x+\delta x)| dx$ and
plot $C_m(\delta x)$ in Fig. \ref{fig_ap1}(b). 
The correlation length of $C_m(\delta x)$ was roughly estimated as 
$1.9$ $\mu$m by fitting $C_m(\delta x)$ with $\exp(-\delta x/1.9)$. 
The correlation length is comparable to the spot size $\sim 2 \mu$m 
of the lensed fiber probe used in this experiment. 

\section{Evaluation of memory capability \label{app_sec2}}
\begin{figure}[h]
\begin{center}
\hspace*{-1cm}
\raisebox{0.0cm}{\includegraphics[width=12cm]{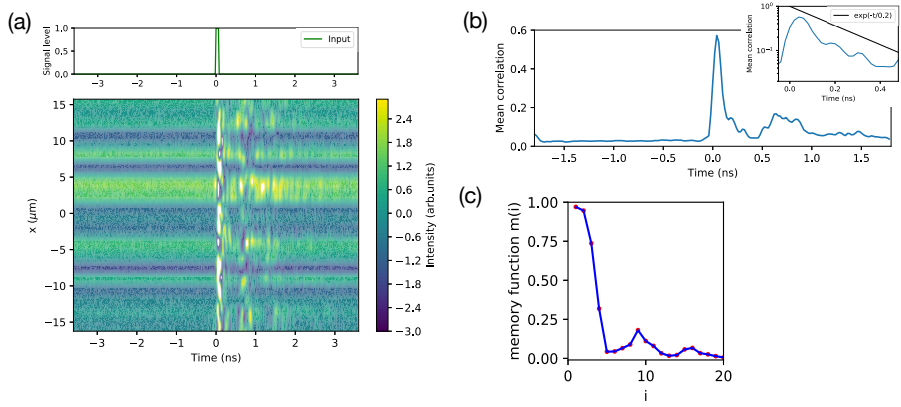}} \\
\end{center}
\caption{\label{fig_ap2}
(a) Input pulse (upper panel) and responses to the pulse, $I(x,t)$ (lower panel). 
(b) The mean absolute value of the correlation $C_m(\Delta t_c)$ as a function of delay time $\Delta t_c$, wherein 
the inset shows a semi-logarithmic plot of $C_m(\Delta t_c)$. 
(c) Result of a memory capacity task. Memory function $m(i)$ is shown as a function of delay step $i$. The input rate was set as $1/\tau = 12.5$ GS/s. 
}
\end{figure}

To evaluate the memory capability of our multimode waveguide device, 
we investigated the device responses to a pulse stimulus.
Figure~\ref{fig_ap2}(a) shows the neural field $I(x,t)$
responding to the input pulse signal $u(t)$ with a width of 0.08 ns. 
The transient responses were observed during a few nanoseconds 
after the pulse stimulus. 
A characteristic time of the responses was evaluated by 
measuring the correlation 
between the input pulse $u(t)$ and the responses $I(x,t)$ as follows:
\begin{align}
C(\Delta t_c,x) =  \dfrac{
\langle [u(t)-\bar{u}_T]
[I(x,t+\Delta t_c)-\bar{I}_{x,T}]
\rangle_T
}
{
\sigma_u\sigma_I},
\end{align}
and computed the mean absolute value of the correlation, 
$C_m(\Delta t_c) = \langle |C(\Delta t_c,x)|\rangle_x$, 
where $\langle \cdot \rangle_x$ denotes the spatial mean. 
The result is plotted in Fig.~\ref{fig_ap2}(b), wherein 
the inset depicts the semi-logarithmic 
plot of $C_m(\Delta t_c)$ in the neighborhood of $t = 0$, 
thereby revealing that $C_m(\Delta t_c)$ exponentially decays 
as the delay time $\Delta t_c$ increases. 
$C_m(\Delta t_c)$ was fitted with an exponential curve, which 
revealed that the correlation time can be 
estimated as 0.2 ns (200 ps). 
Because the correlation time indicates that the information of the input pulse
remains in the device, 
the time is related to the length of the memory.    
The second broad peak from 0.5 ns to 0.8 ns was potentially caused by
the reflection between the ends of the multimode waveguide 
because the time is close to the round 
trip time, $2L/v_g \approx $0.76 ns, where $v_g$ represents the 
estimated group velocity and $L$ denotes the length of the multimode waveguide.   
%

Then, we performed an additional experiment on the memory capability of 
the multimode waveguide device based on a memory capacity task.
In this experiment, the input $u(n)$ 
was set as an identically distributed random 
sequence, uniformly distributed between $[-1,1]$. 
The target values were set as the 
delayed versions of the temporal variation of the input signal, 
i.e., $y_{tag}(n,i) = u(n+1-i)-u(n-i)$ because our device responds well 
to the variation of the input rather than the input itself.
The memory function, $m(i)$, was defined as the normalized linear 
correlation between the target and the predicted output $y(n,i)$;
\begin{align}
 m(i) = \dfrac{
\langle
y_{tag}(n,i)y(n,i)
\rangle_n^2
}
{
\sigma_y^2\sigma_{y_{tag}}^2
},
\end{align}
where $\langle\cdot\rangle_n$ is the mean and $\sigma^2$ denotes the variance. 
Figure~\ref{fig_ap2}(c) shows the memory function $m(i)$ as a function of delay step $i$.
The memory sufficiently decays for $i > 3$. 
In other words, the information of the past input up to $i = 3$ step before
can be remembered.
For the input rate of $1/\tau = 12.5$ GS/s, 
the length of the memory is in a range from $2\tau = 0.16$ ns 
to $3 \tau = 0.24$ ns, 
which is comparable to the time estimated from the pulse-response 
experiment, $t_p \approx 0.2$ ns.  
In the main text, we described the length of the memory time as $0.2$ ns. 

\section{The number of excited guided modes \label{app_sec3}}
\begin{figure}[h]
\begin{center}
\raisebox{0.0cm}{\includegraphics[width=10cm]{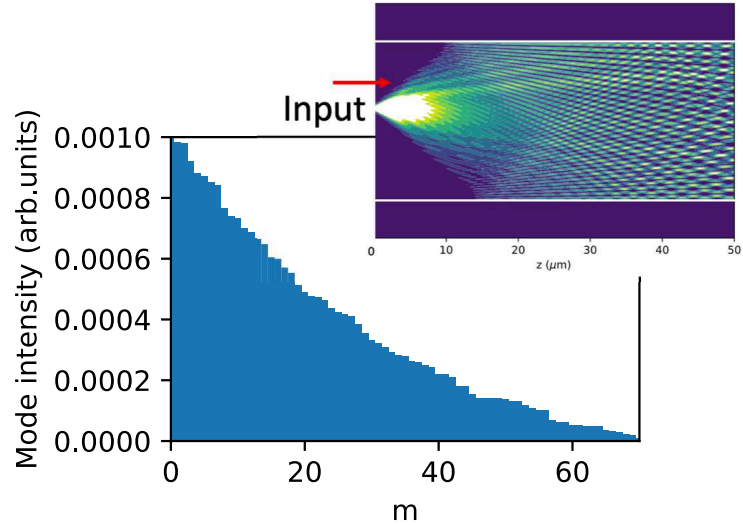}} \\
\end{center}
\caption{\label{fig_ap3}
Numerical results on modal coupling. 
Intensity of the coupling coefficient $|a_m|^2$ of the guided modes 
excited by the light input from the singlemode waveguide. 
The inset shows the intensity distribution around the input port 
to a multimode waveguide, 
where the singlemode waveguide ($z < 0.0$ $\mu$m, not shown in the figure) 
is coupled to the multimode waveguide ($z \ge 0.0$ $\mu$m). 
$z$ denotes coordinates in the propagation direction. 
}
\end{figure}

We computed the guided modes excited by the coupling between 
the 500-nm-wide singlemode waveguide and 25-$\mu$m-multimode waveguide
by solving Eq.~(1) in the main text under the condition 
$E_{in}(x,t) = E_{out}(x,t;0,\omega_0)$.
A lowest-order guided mode of the singlemode waveguide was incident 
from a position slightly deviated from the center of 
the multimode waveguide, 
and it spread owing to the diffraction in the multimode waveguide, 
as shown in the inset of Fig.~\ref{fig_ap3}. 
We numerically confirmed that all the guided modes can be 
excited as a result of the diffraction coupling, 
as shown in Fig.~\ref{fig_ap3}, 
where the intensity of the coupling coefficient 
$|a_m|^2$ is plotted as a function of mode index $m$.  


\begin{thebibliography}{10}
\newcommand{\enquote}[1]{``#1''}

\bibitem{Furber_2016}
S.~Furber, \enquote{Large-scale neuromorphic computing systems,}
  {\it{Journal of Neural Engineering}} \textbf{13}, 051001
  (2016).

\bibitem{6817512}
X.-W. Chen and X.~Lin, \enquote{Big data deep learning: Challenges and
  perspectives,} {\it{IEEE Access}} \textbf{2}, 514--525
  (2014).

\bibitem{arXiv200705558}
N.~C. Thompson, K.~Greenewald, K.~Lee, and G.~F. Manso, \enquote{The
  computational limits of deep learning,}
  {\it{arXiv:2007.05558}} .

\bibitem{Merolla668}
P.~A. Merolla, J.~V. Arthur, R.~Alvarez-Icaza, A.~S. Cassidy, J.~Sawada,
  F.~Akopyan, B.~L. Jackson, N.~Imam, C.~Guo, Y.~Nakamura, B.~Brezzo, I.~Vo,
  S.~K. Esser, R.~Appuswamy, B.~Taba, A.~Amir, M.~D. Flickner, W.~P. Risk,
  R.~Manohar, and D.~S. Modha, \enquote{A million spiking-neuron integrated
  circuit with a scalable communication network and interface,}
  {\it{Science}} \textbf{345}, 668--673 (2014).

\bibitem{Markovic:2020aa}
D.~Markovi{\'c}, A.~Mizrahi, D.~Querlioz, and J.~Grollier, \enquote{Physics for
  neuromorphic computing,} {\it{Nature Reviews Physics}}
  \textbf{2}, 499--510 (2020).

\bibitem{9049105}
W.~Bogaerts and A.~Rahim, \enquote{Programmable photonics: An opportunity for
  an accessible large-volume pic ecosystem,} {\it{IEEE
  Journal of Selected Topics in Quantum Electronics}} \textbf{26}, 1--17
  (2020).

\bibitem{Shastri:2021aa}
B.~J. Shastri, A.~N. Tait, T.~Ferreira~de Lima, W.~H.~P. Pernice, H.~Bhaskaran,
  C.~D. Wright, and P.~R. Prucnal, \enquote{Photonics for artificial
  intelligence and neuromorphic computing,} {\it{Nature
  Photonics}} \textbf{15}, 102--114 (2021).

\bibitem{Wu:2021aa}
C.~Wu, H.~Yu, S.~Lee, R.~Peng, I.~Takeuchi, and M.~Li, \enquote{Programmable
  phase-change metasurfaces on waveguides for multimode photonic convolutional
  neural network,} {\it{Nature Communications}} \textbf{12},
  96 (2021).

\bibitem{TANAKA2019100}
G.~Tanaka, T.~Yamane, J.~B. H{\'e}roux, R.~Nakane, N.~Kanazawa, S.~Takeda,
  H.~Numata, D.~Nakano, and A.~Hirose, \enquote{Recent advances in physical
  reservoir computing: A review,} {\it{Neural Networks}}
  \textbf{115}, 100--123 (2019).

\bibitem{8262616}
R.~Nakane, G.~Tanaka, and A.~Hirose, \enquote{Reservoir computing with spin
  waves excited in a garnet film,} {\it{IEEE Access}}
  \textbf{6}, 4462--4469 (2018).

\bibitem{Du:2017aa}
C.~Du, F.~Cai, M.~A. Zidan, W.~Ma, S.~H. Lee, and W.~D. Lu, \enquote{Reservoir
  computing using dynamic memristors for temporal information processing,}
  {\it{Nature Communications}} \textbf{8}, 2204 (2017).

\bibitem{Appeltant:2011ab}
L.~Appeltant, M.~C. Soriano, G.~Van~der Sande, J.~Danckaert, S.~Massar,
  J.~Dambre, B.~Schrauwen, C.~R. Mirasso, and I.~Fischer, \enquote{Information
  processing using a single dynamical node as complex system,}
  {\it{Nature Communications}} \textbf{2}, 468 (2011).

\bibitem{Tait:2017aa}
A.~N. Tait, T.~F. de~Lima, E.~Zhou, A.~X. Wu, M.~A. Nahmias, B.~J. Shastri, and
  P.~R. Prucnal, \enquote{Neuromorphic photonic networks using silicon photonic
  weight banks,} {\it{Scientific Reports}} \textbf{7}, 7430
  (2017).

\bibitem{PhysRevX.9.021032}
R.~Hamerly, L.~Bernstein, A.~Sludds, M.~Solja\ifmmode \check{c}\else
  \v{c}\fi{}i\ifmmode~\acute{c}\else \'{c}\fi{}, and D.~Englund,
  \enquote{Large-scale optical neural networks based on photoelectric
  multiplication,} {\it{Phys. Rev. X}} \textbf{9}, 021032
  (2019).

\bibitem{Lin1004}
X.~Lin, Y.~Rivenson, N.~T. Yardimci, M.~Veli, Y.~Luo, M.~Jarrahi, and A.~Ozcan,
  \enquote{All-optical machine learning using diffractive deep neural
  networks,} {\it{Science}} \textbf{361}, 1004--1008 (2018).

\bibitem{9064516}
X.~{Sui}, Q.~{Wu}, J.~{Liu}, Q.~{Chen}, and G.~{Gu}, \enquote{A review of
  optical neural networks,} {\it{IEEE Access}} \textbf{8},
  70773--70783 (2020).

\bibitem{Feldmann:2021aa}
J.~Feldmann, N.~Youngblood, M.~Karpov, H.~Gehring, X.~Li, M.~Stappers,
  M.~Le~Gallo, X.~Fu, A.~Lukashchuk, A.~S. Raja, J.~Liu, C.~D. Wright,
  A.~Sebastian, T.~J. Kippenberg, W.~H.~P. Pernice, and H.~Bhaskaran,
  \enquote{Parallel convolutional processing using an integrated photonic
  tensor core,} {\it{Nature}} \textbf{589}, 52--58 (2021).

\bibitem{Xu:2021aa}
X.~Xu, M.~Tan, B.~Corcoran, J.~Wu, A.~Boes, T.~G. Nguyen, S.~T. Chu, B.~E.
  Little, D.~G. Hicks, R.~Morandotti, A.~Mitchell, and D.~J. Moss, \enquote{11
  tops photonic convolutional accelerator for optical neural networks,}
  {\it{Nature}} \textbf{589}, 44--51 (2021).

\bibitem{Kitayama:2019}
K.~Kitayama, M.~Notomi, M.~Naruse, K.~Inoue, S.~Kawakami, and A.~Uchida,
  \enquote{Novel frontier of photonics for data processing---photonic
  accelerator,} {\it{APL Photonics}} \textbf{4}, 090901
  (2019).

\bibitem{Hotchip_9220525}
C.~Ramey, \enquote{Silicon photonics for artificial intelligence acceleration :
  Hotchips 32,} in \emph{2020 IEEE Hot Chips 32 Symposium (HCS),}  (IEEE
  Computer Society, Los Alamitos, CA, USA, 2020), pp. 1--26.

\bibitem{Shen:2017aa}
Y.~Shen, N.~C. Harris, S.~Skirlo, M.~Prabhu, T.~Baehr-Jones, M.~Hochberg,
  X.~Sun, S.~Zhao, H.~Larochelle, D.~Englund, and M.~Solja{\v c}i{\'c},
  \enquote{Deep learning with coherent nanophotonic circuits,}
  {\it{Nature Photonics}} \textbf{11}, 441--446 (2017).

\bibitem{Goodman:78}
J.~W. Goodman, A.~R. Dias, and L.~M. Woody, \enquote{Fully parallel, high-speed
  incoherent optical method for performing discrete fourier transforms,}
  {\it{Opt. Lett.}} \textbf{2}, 1--3 (1978).

\bibitem{Goodman:91}
J.~W. Goodman, \enquote{4 decades of optical information processing,}
  {\it{Opt. Photon. News}} \textbf{2}, 11--15 (1991).

\bibitem{VanderSande:2017aa}
G.~Van~der Sande, D.~Brunner, and M.~C. Soriano, \enquote{Advances in photonic
  reservoir computing,} {\it{Nanophotonics}} \textbf{6},
  561--576 (2017).

\bibitem{Paquot:2012aa}
Y.~Paquot, F.~Duport, A.~Smerieri, J.~Dambre, B.~Schrauwen, M.~Haelterman, and
  S.~Massar, \enquote{Optoelectronic reservoir computing,}
  {\it{Scientific Reports}} \textbf{2}, 287 (2012).

\bibitem{Brunner:2013aa}
D.~Brunner, M.~C. Soriano, C.~R. Mirasso, and I.~Fischer, \enquote{Parallel
  photonic information processing at gigabyte per second data rates using
  transient states,} {\it{Nature Communications}} \textbf{4},
  1364 (2013).

\bibitem{Vandoorne:2014aa}
K.~Vandoorne, P.~Mechet, T.~Van~Vaerenbergh, M.~Fiers, G.~Morthier,
  D.~Verstraeten, B.~Schrauwen, J.~Dambre, and P.~Bienstman,
  \enquote{Experimental demonstration of reservoir computing on a silicon
  photonics chip,} {\it{Nature Communications}} \textbf{5},
  3541 (2014).

\bibitem{Nguimdo:14}
R.~M. Nguimdo, G.~Verschaffelt, J.~Danckaert, and G.~V. der Sande,
  \enquote{Fast photonic information processing using semiconductor lasers with
  delayed optical feedback: Role of phase dynamics,}
  {\it{Opt. Express}} \textbf{22}, 8672--8686 (2014).

\bibitem{Ortin:2015aa}
S.~Ort{\'\i}n, M.~C. Soriano, L.~Pesquera, D.~Brunner, D.~San-Mart{\'\i}n,
  I.~Fischer, C.~R. Mirasso, and J.~M. Guti{\'e}rrez, \enquote{A unified
  framework for reservoir computing and extreme learning machines based on a
  single time-delayed neuron,} {\it{Scientific Reports}}
  \textbf{5}, 14945 (2015).

\bibitem{Vinckier:15}
Q.~Vinckier, F.~Duport, A.~Smerieri, K.~Vandoorne, P.~Bienstman, M.~Haelterman,
  and S.~Massar, \enquote{High-performance photonic reservoir computer based on
  a coherently driven passive cavity,} {\it{Optica}}
  \textbf{2}, 438--446 (2015).

\bibitem{PhysRevX.7.011015}
L.~Larger, A.~Bayl\'on-Fuentes, R.~Martinenghi, V.~S. Udaltsov, Y.~K. Chembo,
  and M.~Jacquot, \enquote{High-speed photonic reservoir computing using a
  time-delay-based architecture: Million words per second classification,}
  {\it{Phys. Rev. X}} \textbf{7}, 011015 (2017).

\bibitem{Takano:18}
K.~Takano, C.~Sugano, M.~Inubushi, K.~Yoshimura, S.~Sunada, K.~Kanno, and
  A.~Uchida, \enquote{Compact reservoir computing with a photonic integrated
  circuit,} {\it{Opt. Express}} \textbf{26}, 29424--29439
  (2018).

\bibitem{Sunada:2019aa}
S.~Sunada and A.~Uchida, \enquote{Photonic reservoir computing based on
  nonlinear wave dynamics at microscale,} {\it{Scientific
  Reports}} \textbf{9}, 19078 (2019).

\bibitem{Uchida:2020JJAP}
A.~Uchida, K.~Kanno, S.~Sunada, and M.~Naruse, \enquote{Reservoir computing and
  decision making using laser dynamics for photonic accelerator,}
  {\it{Jpn. J. Appl. Phys}} \textbf{59}, 040601 (2020).

\bibitem{8807158}
J.~Dong, M.~Rafayelyan, F.~Krzakala, and S.~Gigan, \enquote{Optical reservoir
  computing using multiple light scattering for chaotic systems prediction,}
  {\it{IEEE Journal of Selected Topics in Quantum
  Electronics}} \textbf{26}, 1--12 (2020).

\bibitem{Paudel:20}
U.~Paudel, M.~Luengo-Kovac, J.~Pilawa, T.~J. Shaw, and G.~C. Valley,
  \enquote{Classification of time-domain waveforms using a speckle-based
  optical reservoir computer,} {\it{Opt. Express}}
  \textbf{28}, 1225--1237 (2020).

\bibitem{Sunada:20}
S.~Sunada, K.~Kanno, and A.~Uchida, \enquote{Using multidimensional speckle
  dynamics for high-speed, large-scale, parallel photonic computing,}
  {\it{Opt. Express}} \textbf{28}, 30349--30361 (2020).

\bibitem{PhysRevX.10.041037}
M.~Rafayelyan, J.~Dong, Y.~Tan, F.~Krzakala, and S.~Gigan, \enquote{Large-scale
  optical reservoir computing for spatiotemporal chaotic systems prediction,}
  {\it{Phys. Rev. X}} \textbf{10}, 041037 (2020).

\bibitem{Harkhoe:20}
K.~Harkhoe, G.~Verschaffelt, A.~Katumba, P.~Bienstman, and G.~V. der Sande,
  \enquote{Demonstrating delay-based reservoir computing using a compact
  photonic integrated chip,} {\it{Opt. Express}} \textbf{28},
  3086--3096 (2020).

\bibitem{Nakajima:2021aa}
M.~Nakajima, K.~Tanaka, and T.~Hashimoto, \enquote{Scalable reservoir computing
  on coherent linear photonic processor,} {\it{Communications
  Physics}} \textbf{4}, 20 (2021).

\bibitem{arXiv_Borghi_2021}
M.~Borghi, S.~Biasi, and L.~Pavesi, \enquote{Reservoir computing based on a
  silicon microring and time multiplexing for binary and analog operations,}
  {\it{arXiv:2101.01664}} .

\bibitem{Jaeger78}
H.~Jaeger and H.~Haas, \enquote{Harnessing nonlinearity: Predicting chaotic
  systems and saving energy in wireless communication,}
  {\it{Science}} \textbf{304}, 78--80 (2004).

\bibitem{Maass:2002}
W.~Maass, T.~Natschl{\"a}ger, and H.~Markram, \enquote{{Real-Time Computing
  Without Stable States: A New Framework for Neural Computation Based on
  Perturbations},} {\it{Neural Computation}} \textbf{14},
  2531--2560 (2002).

\bibitem{VERSTRAETEN2007391}
D.~Verstraeten, B.~Schrauwen, M.~D'Haene, and D.~Stroobandt, \enquote{An
  experimental unification of reservoir computing methods,}
  {\it{Neural Networks}} \textbf{20}, 391--403 (2007). Echo
  State Networks and Liquid State Machines.

\bibitem{Antonik:2019aa}
P.~Antonik, N.~Marsal, D.~Brunner, and D.~Rontani, \enquote{Human action
  recognition with a large-scale brain-inspired photonic computer,}
  {\it{Nature Machine Intelligence}} \textbf{1}, 530--537
  (2019).

\bibitem{Dambre:2012aa}
J.~Dambre, D.~Verstraeten, B.~Schrauwen, and S.~Massar, \enquote{Information
  processing capacity of dynamical systems,} {\it{Scientific
  Reports}} \textbf{2}, 514 (2012).

\bibitem{Saleh:1985}
B.~E.~A. Saleh and R.~M. Abdula, \enquote{Optical interference and pulse
  propagation in multimode fibers,} {\it{Fiber and Integrated
  Optics}} \textbf{5}, 161--201 (1985).

\bibitem{Redding:16}
B.~Redding, S.~F. Liew, Y.~Bromberg, R.~Sarma, and H.~Cao,
  \enquote{Evanescently coupled multimode spiral spectrometer,}
  {\it{Optica}} \textbf{3}, 956--962 (2016).

\bibitem{298828}
A.~Weigend and N.~Gershenfeld, \enquote{Results of the time series prediction
  competition at the santa fe institute,} in \emph{IEEE International
  Conference on Neural Networks,}  (1993), pp. 1786--1793 vol.3.

\bibitem{Kuriki:18}
Y.~Kuriki, J.~Nakayama, K.~Takano, and A.~Uchida, \enquote{Impact of input mask
  signals on delay-based photonic reservoir computing with semiconductor
  lasers,} {\it{Opt. Express}} \textbf{26}, 5777--5788
  (2018).

\bibitem{Nahmias:2020}
M.~A. Nahmias, T.~F. de~Lima, A.~N. Tait, H.-T. Peng, B.~J. Shastri, and P.~R.
  Prucnal, \enquote{Photonic multiply-accumulate operations for neural
  networks,} {\it{IEEE Journal of Selected Topics in Quantum
  Electronics}} \textbf{26}, 1--18 (2020).

\bibitem{Angelina:2020}
A.~R. Totovi{\'c}, G.~Dabos, N.~Passalis, A.~Tefas, and N.~Pleros,
  \enquote{Femtojoule per mac neuromorphic photonics: An energy and technology
  roadmap,} {\it{IEEE Journal of Selected Topics in Quantum
  Electronics}} \textbf{26}, 1--15 (2020).

\bibitem{PhysRevApplied.15.034092}
G.~Furuhata, T.~Niiyama, and S.~Sunada, \enquote{Physical deep learning based
  on optimal control of dynamical systems,} {\it{Phys. Rev.
  Applied}} \textbf{15}, 034092 (2021).

\bibitem{Frantz:79}
L.~M. Frantz, A.~A. Sawchuk, and W.~von~der Ohe, \enquote{Optical phase
  measurement in real time,} {\it{Appl. Opt.}} \textbf{18},
  3301--3306 (1979).

\bibitem{Li:09}
G.~Li, \enquote{Recent advances in coherent optical communication,}
  {\it{Adv. Opt. Photon.}} \textbf{1}, 279--307 (2009).

\bibitem{Rawson:80}
E.~G. Rawson, J.~W. Goodman, and R.~E. Norton, \enquote{Frequency dependence of
  modal noise in multimode optical fibers,} {\it{J. Opt. Soc.
  Am.}} \textbf{70}, 968--976 (1980).

\bibitem{LBFGS:2014}
R.~H. Byrd, S.~Hansen, J.~Nocedal, and Y.~Singer, \enquote{A stochastic
  quasi-newton method for large-scale optimization,}
  {\it{arXiv:1401.7020v2}} .

\end{thebibliography}



\end{document}